\documentclass{article}
\usepackage{ijcai25}

\usepackage{times}

\usepackage{soul}
\usepackage{url}
\usepackage[hidelinks]{hyperref}
\usepackage[utf8]{inputenc}
\usepackage[small]{caption}
\usepackage{graphicx}
\usepackage{amsmath}
\usepackage{amsthm}
\usepackage{booktabs}
\usepackage{algorithm}
\usepackage{algorithmic}
\usepackage[switch]{lineno}
\usepackage{multirow}
\usepackage{enumitem}
\usepackage{amssymb}
\usepackage{amsmath}
\usepackage[table,xcdraw]{xcolor}
\usepackage{makecell} 
\usepackage{graphicx}
\usepackage{array,colortbl}  
\usepackage{arydshln}  
\pdfoutput=1
\title{{\scshape LayoutDreamer:}~Physics-guided Layout for Text-to-3D Compositional Scene Generation}
\author{
Yang Zhou
\and
Zongjin He\and
Qixuan Li\And
Chao Wang\footnote{Corresponding author}\\
\affiliations
ShangHai University\\
\emails
\{saber\_mio,
azzi\_counterglew,
liqixuan,
cwang\}@shu.edu.cn
}
\begin{document}
\maketitle
\begin{abstract}
Recently, the field of text-guided 3D scene generation has garnered significant attention. High-quality generation that aligns with physical realism and high controllability is crucial for practical 3D scene applications. However, existing methods face fundamental limitations: (i) difficulty capturing complex relationships between multiple objects described in the text, (ii) inability to generate physically plausible scene layouts, and (iii) lack of controllability and extensibility in compositional scenes. In this paper, we introduce {\scshape LayoutDreamer}, a framework that leverages 3D Gaussian Splatting (3DGS) to facilitate high-quality, physically consistent compositional scene generation guided by text. Specifically, given a text prompt, we convert it into a directed scene graph and adaptively adjust the density and layout of the initial compositional 3D Gaussians. Subsequently, dynamic camera adjustments are made based on the training focal point to ensure entity-level generation quality. Finally, by extracting directed dependencies from the scene graph, we tailor physical and layout energy to ensure both realism and flexibility. Comprehensive experiments demonstrate that {\scshape LayoutDreamer} outperforms other compositional scene generation quality and semantic alignment methods. Specifically, it achieves state-of-the-art (SOTA) performance in the multiple objects generation metric of T$^{3}$Bench.
\end{abstract}

\section{Introduction}

3D models are widely applied in various fields, including autonomous driving, product concept design, gaming, augmented reality (AR), and virtual reality (VR). With rapid advancements in text-to-image models \cite{rombach2022high,saharia2022photorealistic}, text-to-3D generation technology has also made significant progress in generating individual entities \cite{abelson-et-al:scheme,metzer2023latent,poole2022dreamfusion}. However, these models still face challenges in more complex generation tasks, such as creating objects within a contextual surrounding or generating multiple interacting objects. In these cases, they often struggle to accurately capture intricate spatial relationships, leading to inconsistencies and unrealistic outputs. These issues manifest as variations in the appearance from different viewpoints and outputs that fail to adhere to physical constraints. Even generating an interactive 3D asset that integrates with an existing one remains a significant challenge.

Recently, several studies have attempted to extend text-to-3D generation to the creation of compositional 3D scenes. Compositional scene generation refers to creating a coherent layout for a finite set of 3D assets by analyzing their spatial interactions, guided by a detailed scene prompt. Some methods incorporate additional layout information \cite{bai2023componerf,po2024compositional,zhou2024gala3d,cohen2023set}, imposing strict constraints on the spatial arrangement and interactions of objects. However, these methods have inherent {\it limitations: Constraints on flexibility and expansion potential.} These models tend to focus heavily on layout, which restricts the diversity of individual 3D assets and diminishes the consistency between the text input and the generated 3D assets. Another research direction seeks to guide 3D generation using 2D diffusion priors \cite{gao2024graphdreamer,chen2024vp3d,ge2024compgs}. Although these approaches offer greater flexibility in compositional generation and produce high-quality results, they also have the {\it limitation: A single perspective is insufficient to provide 3D consistent cues for compositional interactions.} This leads to significant performance variations across viewpoints, with some perspectives potentially producing unrealistic results.
\begin{figure*}[t]
\centering
\includegraphics[trim=15 80 25 60, clip, width=\linewidth]{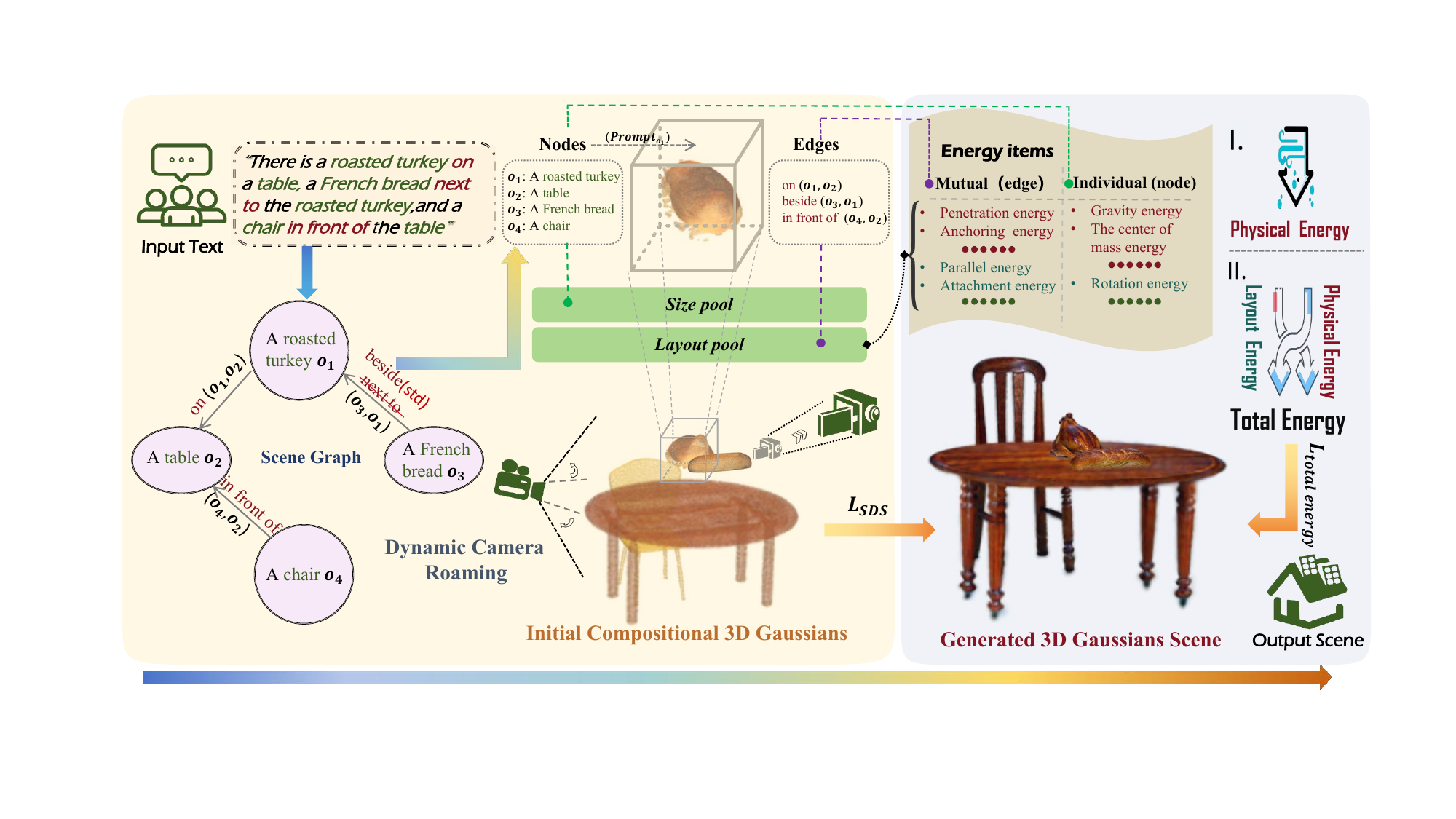}
\caption{\textbf{Overall pipline of \scshape LayoutDreamer.} Given a text prompt, {\scshape LayoutDreamer} convert it into a scene graph, identifying node objects and dependencies. It integrates the size and layout pool to generate initial compositional 3D Gaussians and employs a dynamic camera strategy for entity-level optimization. Energy terms are retrieved from the layout pool based on the scene graph to optimize two-stage layout energy under the principles of physics.}
\label{fig:enter-label_0}
\end{figure*}

To achieve compositional scenes conforming to physical realism, we propose {\scshape LayoutDreamer}, an innovative and scalable framework for generating 3D scenes from intricate text prompts. As shown in Figure~\ref{fig:enter-label_0}, our approach comprises three core components. \textbf{1)} To clarify the interactive relationships within the compositional scene, we present a method specifically developed for initializing compositional 3D Gaussians using scene graphs. Based on the scene graph, the size, density, and position of the initial 3D Gaussians are adaptively adjusted, establishing a disentangled 3D representation. \textbf{2)} To optimize the poses, sizes, positions, and densities of objects in the scene, we propose a dynamic camera roaming strategy that adaptively determines the focal point and focal length during training, ensuring accurate rendering of objects at varying distances and with diverse textures. \textbf{3)} To integrate real-world physical fields, including gravity, mutual penetration, anchoring, and center of mass stability, into the compositional scene, we define a layout energy function by minimizing physical and layout constraints in two stages. This enables a detailed, orderly, and physically consistent arrangement, facilitating the rapid expansion and editing of existing scenes. 

Extensive qualitative and quantitative studies demonstrate that {\scshape LayoutDreamer} can efficiently generate and arrange 3D scenes, ensuring high-fidelity 3D consistency and adherence to physical laws. Our \textbf{contributions} are summarized as follows: 
\begin{enumerate}[label=\arabic*)]
\item To the best of our knowledge, {\scshape LayoutDreamer} is the first text-to-3D compositional scene method by incorporating physical fields, simulating various entity layout scenarios under realistic physical constraints.
\item {\scshape LayoutDreamer} facilitates highly controllable scene editing and expansion by constructing a disentangled representation from a directed scene graph.
\item {\scshape LayoutDreamer} is capable of generating high-fidelity, physics-conforming complex 3D scenes, outperforming SOTA compositional text-to-3D methods.
\end{enumerate}
\section{Related Work}
\subsection{Text-guided 3D Generation}

Early works in text-to-3D generation, such as CLIP-forge \cite{sanghi2022clip}, Dream Fields \cite{jain2022zero}, Text2Mesh \cite{michel2022text2mesh}, CLIP-NeRF \cite{wang2022clip}, and CLIP-mesh \cite{mohammad2022clip} employed CLIP as a guidance mechanism for 3D generation. However, DreamFusion \cite{poole2022dreamfusion} introduced the Score Distillation Sampling (SDS) loss, significantly advancing the quality of 3D models with the aid of 2D diffusion guidance. Magic3D \cite{lin2023magic3d} improved the quality of generated models by employing a two-stage optimization process, progressing from coarse to fine. Fantasia3D \cite{chen2023fantasia3d} prioritized the optimization of geometry and texture in 3D models, while ProlificDreamer \cite{wang2024prolificdreamer} enhanced the diversity of SDS loss and addressed out-of-distribution issues by introducing Variational Score Distillation. Similarly, Score Jacobian Chaining (SJC) \cite{wang2023score} proposed a method for 3D generation using 2D diffusion, leveraging the Perturb-and-Average Scoring (PAAS) technique to iteratively optimize 3D structures. Additionally, other works utilized 3DGS as a 3D representation to achieve rapid and high-fidelity model generation. DreamGaussian \cite{tang2023dreamgaussian} initialized 3D Gaussians by randomly assigning positions within a sphere. However, this approach introduced a bias, favoring spherical symmetry in generated structures. In contrast, methods such as GaussianDreamer \cite{yi2023gaussiandreamer}, GSGEN \cite{chen2024text}, and GaussianDiffusion \cite{li2023gaussiandiffusion} employed pre-trained 3D generation models to initialize 3D Gaussians, offering a more versatile approach.

\subsection{Complex Scene Generation}
Early methods \cite{chang2014learning} for synthesizing 3D scenes used scene graphs to define objects and organize spatial relationships. Giraffe \cite{niemeyer2021giraffe} used compositional NeRF for scene representation, while Set-the-Scene \cite{cohen2023set} developed a style-consistent, disentangled NeRF-based framework for scene generation. Text2Room \cite{hollein2023text2room} and Text2NeRF \cite{zhang2024text2nerf} generated 2D views from text and extrapolated these views to construct 3D scenes but struggled to maintain scene coherence. VP3D \cite{chen2024vp3d} and CompGS \cite{ge2024compgs} achieved compositional 3D generation through layout guidance from 2D views. CG3D \cite{vilesov2023cg3d} incorporated gravity and contact constraints during compositional generation to produce physically realistic outcomes. 

With the rise of large language models (LLMs), new inspirations for scene layout have emerged. Methods such as SceneCraft \cite{kumaran2023scenecraft}, Holodeck \cite{yang2024holodeck} and LayoutGPT \cite{feng2024layoutgpt} utilized LLMs or vision-language models (VLMs) to generate complex 3D scenes from the textual descriptions. Nonetheless, due to the hallucination issues inherent in large models, layout confusion can arise in intricate spatial environments. Gala3D \cite{zhou2024gala3d} utilized coarse layout priors from LLMs and refined the layout through optimization to achieve more structured and coherent scene arrangements. 

\section{Method}
\subsection{Overview}
As shown in Figure~\ref{fig:enter-label_0}, given a text prompt $T_p$ to generate a scene $O =\{ o_i \}_{i=1}^M$ with $M$ objects, we begin by constructing a scene graph $G(O)$ using methods for entity and relationship extraction. To initialize the 3D entities, we generate point clouds using Shap-E \cite{jun2023shap}, which are then converted into 3D Gaussians. We introduce a density adjustment method based on the size pool and a chain-based positioning method utilizing layout pools to optimize objects' size, density, and position. Next, we employ a decomposed optimization strategy to iteratively train and refine the scene, performing $M$ camera roams with an adaptive strategy to optimize the generation of entities (Section \ref{sec:3.3}). Following this, we use the scene graph to derive scene-guided configurations, allowing us to customize the scene's physical and layout constraints. These constraints are then optimized under a dynamic, hierarchical energy function, ensuring a neat and logically consistent arrangement of objects (Section \ref{sec:3.4}).

\subsection{Scene Graph-guided Initial 3D Gaussians}
\label{sec:3.2}
Given a user text prompt $T_p$, a directed scene graph is constructed by parsing the objects and spatial dependencies described in the text. In this graph, object entities are represented as nodes and various relationships are mapped to standardized forms, represented as directed edges (e.g., `on' and `upon' are mapped to the standard relation `on').

\subsubsection{Scale-aware Density Adjustment}
To ensure that the generated initial 3D Gaussians volumes adhere to real-world dimensional standards, we design a size pool for the nodes in the scene graph. The size pool comprises object categories, size levels, and corresponding values. After two rounds of semantic similarity matching, each object is assigned to a specific size pool, and its standard size value $S_i$ (where $i \in M$) is determined. Considering both the standard size and the current object's size, we apply a scale-aware density adjustment technique to ensure that after scaling, 3D Gaussians maintain consistent density while preserving essential geometric details. Specifically, when the standard size exceeds the current size, we perform interpolation based on the volume ratio before and after scaling to increase the density of 3D Gaussians. Conversely, when the standard size is smaller than the current size, we use a combined method of voxel grid downsampling and geometric feature sampling to reduce the number of 3D Gaussians for smaller objects. This approach minimizes training overhead while retaining essential geometric feature points.
\subsubsection{Chain-based Position Initialization}
In complex scene interactions, an object's spatial position is determined by its interaction relationships and the positions of other involved objects. To obtain a coarse scene layout, we introduce a layout pool for the directed edges in the scene graph. The pool contains the standard offset $\Delta P(r_k)$ for each standard dependency relationship $r_k$, along with energy term weights used during layout training (Section~\ref{sec:3.4}). Each object is processed according to topological sorting, with all incoming spatial dependencies aggregated for updates. For each object $o_i$, its position $P(o_i)$ is determined by all incoming relationships: 
\begin{equation}
    P(o_i) = \sum_{k} (P(s_k) + \Delta P(r_k)) ,
\end{equation}
\begin{equation}
    \Delta P(r_k) = [\Delta x(r_k), \Delta y(r_k), \Delta z(r_k), \Delta d(r_k)] , 
\end{equation}
where $s_k$ is the dependent object and $\Delta x(r_k)$, $\Delta y(r_k)$, $\Delta z(r_k)$ represent the standard directional offsets for relationship $r_k$. $\Delta d(r_k)$ denotes the distance scaling offset. To differentiate each entity, we assign an independent feature label $L =\{ l_i \}_{i=1}^M$ and incorporate it with the information gathered from the size and layout pool into a scene-guided configuration $\text{Configs}(o_i) = \{ o_i, l_i, P(o_i), S_i \}$.

\subsection{Dynamic Camera Roaming Driven by Training Focus}
\label{sec:3.3}
The camera configuration plays a crucial role in SDS \cite{poole2022dreamfusion}, especially when capturing scenes with occlusion relationships. With static camera configuration facing the origin, issues such as incomplete information from objects at varying positions may arise due to perspective limitations. Additionally, significant size differences between objects can lead to challenges: larger objects may experience internal Janus problems during SDS optimization, while smaller objects may lack detailed texture information. Therefore, we design a dynamic camera roaming strategy driven by the training focus. During entity-level training optimization, the label $l_i$, size $S_i$, and position information $P(o_i)$ of the current object are directly retrieved from the scene-guided configuration. We unfreeze only the parameter groups corresponding to the current label for entity training, while the camera tracks the entity and adjusts its position based on the object’s location. The camera's orientation $d_i$ is recalculated towards the object's center after the adjustment. By evaluating the ratio of the object's actual size to the camera-defined standard size, we can determine a distance adjustment factor $\alpha_i$, which is used to adjust the camera depth. The final camera position is given by:
\begin{equation}
    C' = C + P(o_i) - a_i \cdot d_i \quad \text{with} \quad d_i = \frac{P(o_i) - C}{\|P(o_i) - C\|} ,
\end{equation}
where $C'$ is the adjusted camera position, and $C$ is the original camera position. By adjusting both the camera's translation and depth, objects within the field of view are rendered optimally.

Irregular object edges may lead to layout complexity and cause interpenetration issues. To mitigate this, we encourage the transmittance of the foreground to approach either $0$ or $1$. This technique facilitates the removal of floating objects and corrects 3D Gaussians whose edges are significantly impacted by variations in 2D diffusion guidance results, inspired by \cite{fridovich2022plenoxels,shriram2024realmdreamer}. In a scene containing $M$ objects, disentangled scene generation is achieved by training each object separately, ensuring high-quality, 3D-consistent, and well-separated objects. The total loss during the entity generation phase is expressed as:
\begin{equation}
    L = \sum_{i=1}^{M} (L_{\text{SDS}}(i) + \lambda_o L_o(i)\big) ,
\end{equation}
where $\lambda$ is a hyperparameter controlling the contribution of the opacity loss.

\subsection{Physical Field Integration through Layout Energy Function}
\label{sec:3.4}

By utilizing the edge relationships in the scene graph, we stabilize the scene layout by minimizing the total energy. We define methods for both physical and layout energy, then integrate the layout pool to derive the corresponding energy terms and their respective weights, ultimately resulting in the final total energy.
\subsubsection{Design of Energy Models Reflecting Physical Reality}

To ensure the compositional generation process adheres to the principles of physical reality, we simulate various physical layout conditions, including gravity, the influence of centroid on positioning, and the non-penetration and mutual anchoring of objects. Simultaneously, other layout energy terms are introduced to refine the spatial relationships.

\textbf{Gravity energy term. }To stabilize objects under gravity, we define the following bounding boxes for each entity to efficiently evaluate their direction and posture within the explicit 3DGS representation. Specifically, we set $z=0$ as the ground plane. The gravity energy term stabilizes the object’s position by minimizing the height deviation at the bottom of the following bounding box, ensuring that objects settle onto the ground. This term is expressed as:
\begin{equation}
    \mathbf{E}_g^{(i)} = \text{mean}(z')^2 + \lambda \cdot \max(0, -\min(z')) ,
\end{equation}
where $z'$ is the height of the bottom vertices of the following bounding boxes. 

\textbf{Penetration energy term. }To ensure proper contact between objects and prevent mutual penetration in the constraint optimization problem of a multi-object compositional system, we draw inspiration from CG3D \cite{vilesov2023cg3d} to define the penetration energy term $\mathbf E_p^{(i)}$. For a Gaussian with mean $\mu_i$ in object $O_2$, centered at $q_2$, and a Gaussian with mean $\mu_j$ in $O_1$, which is the closest to $O_2$, a penalty based on the negative cosine is applied to enforce the angle $\phi_i$ between vectors $v_1 = \mu_i - q_2$ and $v_2 = \mu_j - \mu_i$ to be acute, thus preventing penetration between the two objects. The penetration energy term is:
\begin{equation}
    \mathbf{E}_p^{(i)} = \frac{k}{N} \sum_{i=1}^{N} \max\left(0, -\cos(\phi_i)\right),
\end{equation}
where $k$ is the repulsive strength coefficient and $N$ is the number of Gaussians in $O_2$.

\textbf{Anchor energy term. }Considering special scenarios, such as hook-like relationships, we design an anchor energy term activated when the penetration energy term is triggered, ensuring that the anchor points do not experience undesirable shifts or aggregation. When two objects come into contact, each has an associated anchor point $A_i^{(l)}$ in its local coordinate, transformed into world coordinates $A_i^{(w)}$ during layout training. The anchor energy term $\mathbf E_a^{(i,j)}$ penalizes deviations between actual and expected anchor point distances, modeled as elastic potential energy:
\begin{equation}
    \mathbf{E}_a^{(i,j)} = \frac{1}{2} k \left( \| {A}_2^\text{w} - {A}_1^\text{w} \| - {d} \right)^2 ,
\end{equation}
where $d$ denotes the expected distance between the anchor points, and $k$ is the spring constant hyperparameter, controlling the intensity of the anchor constraint. 

\textbf{Other energy terms. }Proper positioning of an object’s centroid is essential for maintaining system stability and physical plausibility.  Therefore, we introduce a centroid energy term that minimizes the vertical displacement of object centroids. Additionally, the layout energy function enforces adherence to physical laws while preserving semantic coherence and visual appeal. To complement the centroid energy term, we propose an alignment energy term, which minimizes the directional differences between the nearest principal axes of two objects across different dimensions, quantifying deviations between the centers of mass of the objects. By reducing the centroid difference along the target direction, the alignment energy term enhances spatial orderliness and promotes logical object arrangement. For optimizing object distances, we implement a proximity energy term that limits sparsity by calculating the distance between the closest points of distant objects. This term ensures that objects maintain an appropriate level of spatial coherence while preventing excessive gaps in the scene. When an anchor energy term is required, the attachment energy term enforces the ideal distance between the nearest points of two interacting objects, which helps maintain stable relationships in anchor-reliant scenarios. While the centroid energy term optimizes object positions globally, it may sometimes induce unnatural rotations, compromising physical realism. To address this, we include a rotation energy term that restricts the maximum allowable rotation angle, ensuring the layout remains physically consistent and visually plausible.

\subsubsection{Optimization of Compositional Scene}

To optimize the compositional scene layout, we freeze the other parameters of the 3D Gaussians parameter groups and focus solely on training the translation and rotation parameters. Within the defined energy constraints, the optimization includes mutual energy terms (e.g., the penetration energy) and individual energy terms (e.g., the gravity energy). By traversing the nodes and directed edges of the scene graph, we assign energy terms and weights to each node systematically. The total constrained energy function is divided into the layout and physical energy functions, with priority given to the latter. Together, these energy terms define the overall energy functions:

\begin{small}
    \begin{equation}
        \mathbf{E}_k = \sum_{(i,j) \in \epsilon} \sum_{k \in p \text{ or } \in l} w_k^{(i,j)} \mathbf{E}_k^{(i,j)} + \sum_{i \in N} \sum_{k \in p \text{ or } \in l} w_k^{(i)} \mathbf{E}_k^{(i)} ,
    \end{equation}    
\end{small}
where $\epsilon$ and $N$ represent all the edges and nodes in the scene graph. $\mathbf{E}_p^{(i,j)}$ and $\mathbf{E}_l^{(i,j)}$ denote the mutual physical and layout energy terms between two connected entities, while $\mathbf{E}_p^{(i)}$ and $\mathbf{E}_l^{(i)}$ refer to the individual physical and layout energy terms for each node. Additionally, $w_k$ represents the weight associated with each energy term.

To achieve an optimal configuration that satisfies physical constraints during scene layout training, we propose a two-phase hierarchical energy minimization method. The total energy function, encompassing both physical and layout energy terms, is expressed as:
\begin{equation}
\mathbf{E}^{(t)} = \lambda_p^{(t)} \hat{\mathbf{E}}_p + \lambda_l^{(t)} \hat{\mathbf{E}}_l,
\end{equation}
where $\hat{\mathbf{E}}_p$ and $\hat{\mathbf{E}}_l$ represent the physical and layout energy functions after $L2$ regularization, ensuring that energy terms of different magnitudes are scaled to a unified order of magnitude. $\lambda_p^{(t)}$ and $\lambda_l^{(t)}$ are the physical and layout constraint weights at step $t$, respectively.
\begin{equation}    
\hat{\mathbf{E}}_p = \frac{\mathbf{E}_p}{\sqrt{\|\mathbf{E}_p\|_2^2 + \|\mathbf{E}_l\|_2^2}},
\hat{\mathbf{E}}_l = \frac{\mathbf{E}_l}{\sqrt{\|\mathbf{E}_p\|_2^2 + \|\mathbf{E}_l\|_2^2}},
\end{equation}
\begin{equation}
\begin{aligned}
\lambda_p^{(t)} &=
\begin{cases} 
1, & t < x, \\
1 - \frac{\beta}{2} \left( 1 - \cos\left(\pi \frac{t - x}{T - x}\right) \right), & x \leq t \leq T ,
\end{cases} \\
\lambda_l^{(t)} &=
\begin{cases} 
0, & t < x, \\
1 - \lambda_p^{(t)}, & x \leq t \leq T ,
\end{cases}
\end{aligned}
\label{eq:lamada}
\end{equation}
In Equation (\ref{eq:lamada}), $x$ denotes the number of steps required for the physical energy to reach its threshold, and $T$ is the total number of training steps. $\beta$ controls the amplitude of the physical energy weight (where $0<\beta<1$). Initially, training emphasizes physical energy constraints until the physical energy falls below the threshold at step $x$. Afterward, a two-phase joint training process optimizes both physical and layout energy. Cosine functions alternate the weights of physical and layout energy, reducing the risk of getting trapped in local minima of the physical energy function. Meanwhile, the physical energy weight is gradually increased towards the end of training to ensure the layout aligns with physical reality.

\section{Experiments}
\subsection{Implementations Details}
{\scshape LayoutDreamer} is implemented using PyTorch and is built upon ThreeStudio \cite{liu2023threestudio}. For the complex prompts in {T$^{3}$Bench}, we use the 8B Llama3 model to extract the subjects and relationships from the text. We employ GaussianDreamer \cite{yi2023gaussiandreamer} as the multi-view diffusion model. The process requires 2000 iterations to train a single object. However, for layout optimization in a scene containing three objects and 15 energy constraints, convergence is achieved within just 300 steps. Our experiments can be completed using a single RTX 3090 GPU with 43G memory. The average total generation time for a scene with $M$ objects is $21 \times M + 2 \times C_M^2$ minutes, where generating a single object takes approximately 20 minutes. Additionally, each pair of mutual energy terms requires about 2 minutes for computation, while calculating the total energy for an individual object’s energy term takes approximately 1 minute.
\begin{figure}[t]
    \centering
    \includegraphics[width=\linewidth]{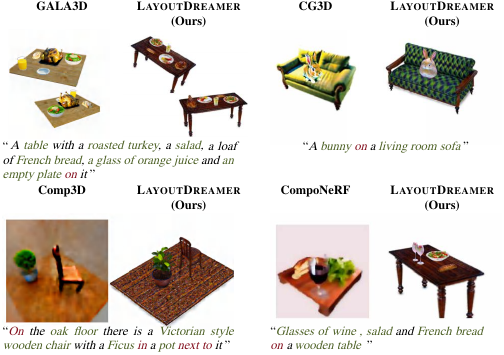}
    \caption{\small \textbf{Comparisons with closed-source compositional text-to-3D methods. }{\scshape LayoutDreamer} emphasizes the layout based on an understanding of physical principles.}
    \label{fig:CLosed}
\end{figure}
\setlength{\dashlinedash}{2pt} 
\setlength{\dashlinegap}{2pt}  
\setlength{\arrayrulewidth}{1.2pt} 
\arrayrulecolor{gray} 
\definecolor{obj}{HTML}{3B5F21}
\definecolor{req}{HTML}{851321}
\definecolor{mygreen}{HTML}{249087}
\begin{figure*}
    \centering
    \includegraphics[width=0.93\linewidth]{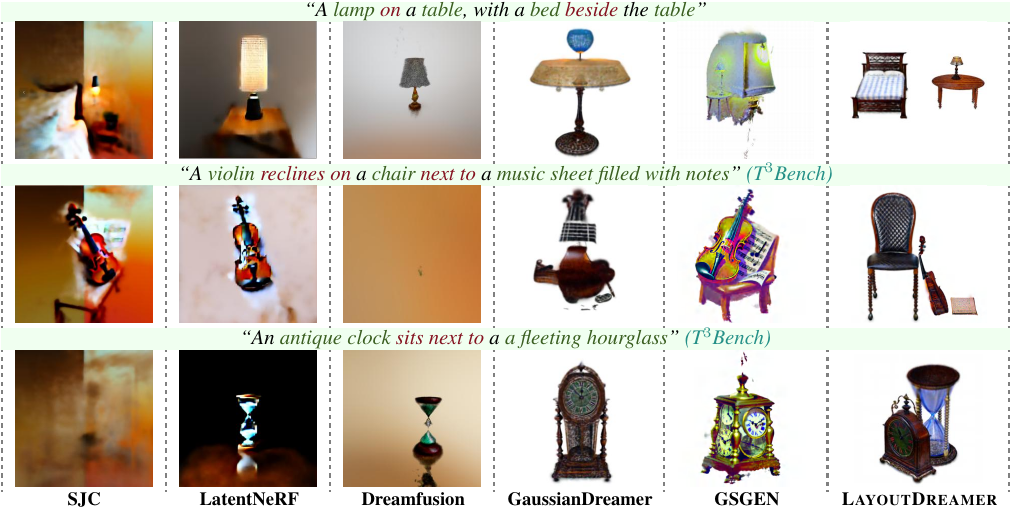}
    \caption{\small  \textbf{Qualitative comparisons between {\scshape LayoutDreamer} with other text-to-3D methods.} The prompts are derived from the standard compositional scene prompts and the multiple objects tracking prompt set provided by T$^{3}$Bench. {\scshape LayoutDreamer} generates disentangled scenes using the same text prompts, with a focus on layout informed by physical principles.}
    \label{fig:mainQC}
\end{figure*}
\subsection{Comparisons with Other Methods}
To validate the effectiveness of our method, we evaluate generation quality and text alignment using the \text{T$^{3}$Bench} \cite{he2023t} evaluation criteria, which provide a comprehensive set of metrics for text-to-3D generation, particularly focusing on multiple objects compositional generation.
\paragraph{Qualitative Comparison.} We compare our method with recent works in composition scene generation that use layouts to guide 3D scene generation. Since most of these works are not open-source, we directly reference results presented in their papers and use identical prompts to generate comparable 3D scenes. As shown in Figure~\ref{fig:CLosed}, both Comp3D \cite{po2024compositional} and CompoNeRF \cite{bai2023componerf} suffer from scene blurring, while CG3D \cite{vilesov2023cg3d} offers a reasonable layout but lacks spatial orderliness. GALA3D \cite{zhou2024gala3d} achieves good decoupled generation; our method excels by producing 3D assets with superior texture detail and complete recognition of entity prompts. Furthermore, we compare {\scshape LayoutDreamer} with several open-source methods for text-to-3D generation, including SJC \cite{wang2023score}, LatentNeRF \cite{metzer2023latent}, Dreamfusion \cite{poole2022dreamfusion} and methods that use point clouds for 3D Gaussians initialization, such as Gaussiandreamer \cite{yi2023gaussiandreamer} and GSGEN \cite{chen2024text}. As shown in Figure~\ref{fig:mainQC}, {\scshape LayoutDreamer} generates physically realistic, high-quality 3D scenes, surpassing other methods in terms of geometry, color, and texture.
\begin{table}[t]
    \small
    \setlength{\tabcolsep}{4pt}
    \renewcommand{\arraystretch}{1.0}
    \centering
    \begin{tabular}{lrrr}
        \toprule
        \multirow{2}{*}{\textbf{Method}} & \multicolumn{3}{c}{\textbf{T$^{3}$Bench (Multiple Objects)}}  \\
        & \textbf{Quality$\uparrow$} & \textbf{Alignment$\uparrow$} & \textbf{Average$\uparrow$} \\
        \midrule
        \small DreamFusion & 17.3 & 14.8 & 16.1 \\
        \small SJC & 17.7 & 5.8 & 11.7 \\
        \small Latent-NeRF & 21.7 & 19.5 & 20.6 \\
        \small DreamGaussian & 12.3 & 9.5 & 10.9 \\
        \small ProlificDreamer & \cellcolor[HTML]{FFF3CA}45.7 & 25.8 & \cellcolor[HTML]{FFF3CA}35.8 \\
        \small MVDream & 39.0 & \cellcolor[HTML]{FFF3CA}28.5 & 33.8 \\
        \small Magic3D & 26.6 & 24.8 & 25.7 \\
        \small VP3D & \cellcolor[HTML]{FCE4D3}49.1 & \cellcolor[HTML]{FCE4D3}31.5 & \cellcolor[HTML]{FCE4D3}40.3 \\
        \midrule
        \small \textbf{{\scshape LayoutDreamer}} & \cellcolor[HTML]{FADADE}(\textcolor{red}{+7.5}) \textbf{56.6} & \cellcolor[HTML]{FADADE}(\textcolor{red}{+0.3}) \textbf{31.8} & \cellcolor[HTML]{FADADE}(\textcolor{red}{+3.9}) \textbf{44.2} \\
        \small \textbf{{\scshape LayoutDreamer}} \\(\textbf{{\scriptsize prompt\_scene only})} & (\textcolor{red}{+18.0}) \textbf{67.1} & (\textcolor{red}{+4.3}) \textbf{35.8} & (\textcolor{red}{+11.2}) \textbf{51.5} \\
        \bottomrule
    \end{tabular}
    \caption{Quantitative comparison on \text{T$^{3}$Bench} with other methods}
    \label{tab:t3bench}
\end{table}
\paragraph{Quantitative Comparison.} In Table~\ref{tab:t3bench}, we benchmark representative models for text-to-3D generation on \text{T$^{3}$Bench}, comparing the results with the related work VP3D \cite{chen2024vp3d}, with a focus on multiple objects generation. The results show that {\scshape LayoutDreamer} achieves the highest quality and text alignment scores. Compared to methods specifically designed for scene generation, {\scshape LayoutDreamer} demonstrates significant advantages in both metrics with its full disentanglement of scene generation and fine-tuned layout optimization. Due to the large number of prompts in \text{T$^{3}$Bench} for multiple objects generation that are not related to compositional scene generation, we use GPT-4o to filter a set of 64 prompts suitable for small scene generation (prompt\_scene) and compare them with the results from the {\scshape LayoutDreamer} method using all prompts. The full prompt set typically includes interaction elements, such as human actions and poses, which are unsuitable for 3D generation using an entity placement-based approach. As shown in the last row of Table \ref{tab:t3bench} shows that the generation quality and text alignment scores evaluated with the scene prompt set from {T$^{3}$Bench} significantly exceed those obtained using the full multiple objects prompt set, showcasing its immense potential in compositional scene tasks.

\subsection{Ablation Studies}

To validate the effectiveness of the key components of {\scshape LayoutDreamer}, we design ablation experiments for compositional 3D Gaussians initialization (CGS Init.), the dynamic camera roaming strategy (DCR), and layout energy constraints (LEC).
\setlength{\dashlinedash}{2pt} 
\setlength{\dashlinegap}{2pt}  
\setlength{\arrayrulewidth}{1.2pt} 
\arrayrulecolor{gray} 
\begin{figure}[t]
    \centering
    \includegraphics[width=0.95\linewidth]{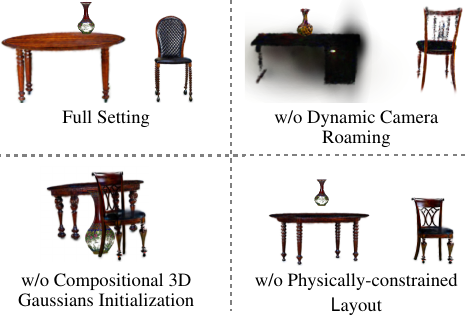}
    \caption{\textbf{Visual results of the ablation studies.} Experiments validate the effectiveness of the three core modules, highlighting the critical roles of scene optimization from coarse to fine layout and entity optimization based on an dynamic camera roaming strategy in compositional scene generation.}
    \label{fig:Abo}
\end{figure}
\begin{figure}[b]
    \centering
    \includegraphics[width=0.95\linewidth]{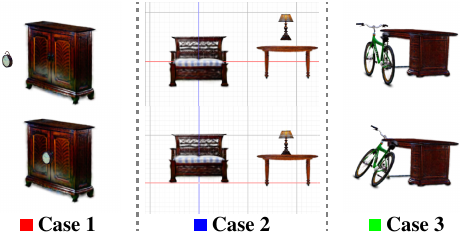}
    \caption{
        \textbf{Validation cases of physical energy terms.} The text prompts for \textcolor{red} {\rule{6pt}{6pt}} Case 1, \textcolor{blue}{\rule{6pt}{6pt}} Case 2, \textcolor{green}{\rule{6pt}{6pt}} Case 3 are as follows: ``a clock hangs on a moldy cabinet", ``a lamp on a table, with a bed beside the table" and ``a bicycle leans against a table".
    }
    \label{fig:Phy}
\end{figure}
\begin{figure}[t]
    \centering
    \includegraphics[width=0.85\linewidth]{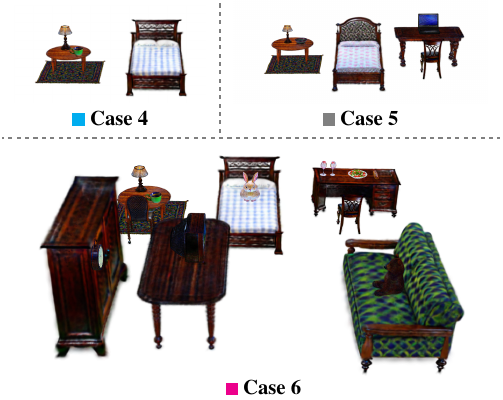}
    \caption{\textbf{Editable and expandable scenes with controllable text prompt.} \textcolor{gray}{\rule{6pt}{6pt}} Case 5 building upon \textcolor{cyan}{\rule{6pt}{6pt}} Case 4 is expanded with the text prompt: ``a table next to the bed, a chair in front of the table, and a computer on the table", enable scene editing, including deletion, movement, and style updates. \textcolor{magenta}{\rule{6pt}{6pt}} Case 6 demonstrates {\scshape LayoutDreamer} is capable of achieving scene expansion at a 
    larger scale.}
    \label{fig:EDIT}
\end{figure}
\paragraph{Compositional 3D Gaussians Initialization.} To achieve the initial compositional 3D Gaussians expression, we employ a scale-aware density adjustment combined with a chain-based position initialization. However, we directly generate the initial 3D Gaussians using point cloud priors and do not use chain-based position initialization to impose rough layout information on the 3D Gaussians entities. As shown in Figure~\ref{fig:Abo}, the 3D Gaussians entities are initialized with unrealistic sizes. The coarse layout, characterized by mutual penetration, leads to confusion in the layout optimization process.

\paragraph{Static Random Camera Capture.} In the forward rendering process of 3DGS, we do not employ a dynamic camera roaming strategy to adjust the camera's intrinsic and extrinsic parameters. Instead, we use a default camera pose with a radius ranging from 1.5 to 4.0, an azimuth angle from -180 to 180 degrees, and an elevation angle from -10 to 60 degrees for scene capture. As shown in Table~\ref{tab:Abo}, the results in a substantial decrease in CLIP scores. Moreover, the visualization in Figure~\ref{fig:Abo} confirms that the initial camera pose fails to adequately capture individual 3D Gaussians entities.
\begin{table}[ht]
\centering
\resizebox{\linewidth}{!}{
\begin{tabular}{lrrrr}
    \toprule
    \textbf{Method} & \small -w/o CGS Init. & \small -w/o DCR & \small -w/o LEC & \textbf{Full Setting} \\
    \midrule
    \textbf{CLIP$\uparrow$} & 28.8 & 25.8 & 33.2 & \textbf{36.3} \\
    \bottomrule
\end{tabular}
}
\caption{Quantitative ablation study of the three key components in {\scshape LayoutDreamer} using CLIP scores.}
\label{tab:Abo}
\end{table}

\paragraph{Physically-constrained Layouts.} Since the layout energy constraints involve numerous energy terms, we design experiments that focus on the specific characteristics of each physical constraint to validate their effectiveness. In Case 1, the clock, which lacks attachment and anchor energy terms, fails to hang from the moldy cabinet. In Case 2, objects without penetration and gravity energy terms float in the air due to the coarse initialization of the 3D Gaussians layout. In Case 3, by introducing centroid, penetration, and rotation energy terms, the bicycle naturally leans against the table and maintains balance.

\subsection{Scalable Disentangled Scene Layout} {\scshape LayoutDreamer} is compatible with all 3DGS representations, offering enhanced control over disentangled scenes by designing scene-guided configurations for each entity. As illustrated in Figure~\ref{fig:EDIT}, {\scshape LayoutDreamer} allows for efficient removal, movement, and regeneration of objects, providing precise control over the scene composition. Additionally, it supports the dynamic combination and rearrangement of 3D Gaussians scene representations alongside scene-guided configuration, enabling seamless dynamic expansion. This rapid scene editing and incremental expansion make {\scshape LayoutDreamer} well-suited for practical real-world applications requiring adaptive and scalable 3D asset creation.

\section{Conclusion}

We introduce {\scshape LayoutDreamer}, a framework for rapidly generating physically realistic and well-structured 3D scenes using text prompts, demonstrating high-quality scene generation and consistency. {\scshape LayoutDreamer} provides a reasonable initialization approach for the domain of compositional 3D Gaussians scene generation. By converting the text into a scene graph, the generated scene achieves an organized layout based on spatial interactions and physical constraints within 15 minutes, allowing users to conveniently and efficiently edit and expand disentangled scenes. Experimental results show that {\scshape LayoutDreamer} outperforms existing methods in text-to-scene generation, effectively handling intricate text to create dynamic interactions among multiple objects while adhering to real-world physical principles.
\bibliographystyle{named}
\bibliography{ijcai25}

\end{document}